\documentclass[letterpaper, 10 pt,conference]{IEEEtran}
\usepackage{cite}
\usepackage{amsmath,amssymb,amsfonts}
\usepackage{algorithmic}
\usepackage{graphicx}
\usepackage{textcomp}
\usepackage{soul}
\usepackage{xcolor}
\usepackage{siunitx}
\usepackage{svg}
\usepackage{verbatim}
\usepackage{subfig}
\usepackage[hidelinks]{hyperref}
\usepackage{flushend}
\usepackage{tikz,xcolor}
\definecolor{lime}{HTML}{A6CE39}
\DeclareRobustCommand{\orcidicon}{
	\begin{tikzpicture}
	\draw[lime, fill=lime] (0,0) 
	circle [radius=0.16] 
	node[white] {{\fontfamily{qag}\selectfont \tiny ID}};
	\draw[white, fill=white] (-0.0625,0.095) 
	circle [radius=0.007];
	\end{tikzpicture}
	\hspace{-2mm}
}

\usepackage{booktabs}
\def\BibTeX{{\rm B\kern-.05em{\sc i\kern-.025em b}\kern-.08em
    T\kern-.1667em\lower.7ex\hbox{E}\kern-.125emX}}
\begin{document}

\foreach \x in {A, ..., Z}{%
	\expandafter\xdef\csname orcid\x\endcsname{\noexpand\href{https://orcid.org/\csname orcidauthor\x\endcsname}{\noexpand\orcidicon}}
}

\newcommand{\orcidauthorA}{0000-0003-1615-3485} 
\newcommand{\orcidauthorB}{0000-0003-1760-3871} 
\newcommand{\orcidauthorC}{0000-0002-7737-8664} 
\newcommand{\orcidauthorD}{0000-0002-9897-4748} 

\title{Object recognition for robotics from tactile time series data utilising different neural network architectures}

\author{
\IEEEauthorblockN{Wolfgang B\"{o}ttcher\orcidA{}}
\IEEEauthorblockA{\textit{ETH Zurich} \\
Zurich, Switzerland\\
wboettcher@student.ethz.ch}
\and
\IEEEauthorblockN{Pedro Machado\orcidB{}}
\IEEEauthorblockA{\textit{Computational Neurosciences and Cognitive Robotics Group}\\
\textit{School of Science and Technology}\\
\textit{Nottingham Trent University}\\
Nottingham, UK\\
pedro.baptistamachado@ntu.ac.uk}
\and
\IEEEauthorblockN{Nikesh Lama\orcidC{}}
\IEEEauthorblockA{\textit{Computational Neurosciences and Cognitive Robotics Group}\\
\textit{School of Science and Technology}\\
\textit{Nottingham Trent University}\\
Nottingham, UK\\
nikesh.lama02@ntu.ac.uk}
\and
\IEEEauthorblockN{T.M. McGinnity\orcidD{}}
\IEEEauthorblockA{\textit{Computational Neurosciences and Cognitive Robotics Group}\\
\textit{School of Science and Technology}\\
\textit{Nottingham Trent University}\\
Nottingham, UK\\
martin.mcginnity@ntu.ac.uk\\
\textit{Intelligent Systems Research Centre}\\
\textit{Ulster University}\\
Northern Ireland, UK\\
tm.mcginnity@ulster.ac.uk}}

\maketitle

\begin{abstract}
Robots need to exploit high-quality information on grasped objects to interact with the physical environment. Haptic data can therefore be used for supplementing the visual modality. This paper investigates the use of Convolutional Neural Networks (CNN) and Long-Short Term Memory (LSTM) neural network architectures for object classification on Spatio-temporal tactile grasping data. Furthermore, we compared these methods using data from two different fingertip sensors (namely the BioTac SP and WTS-FT) in the same physical setup, allowing for a realistic comparison across methods and sensors for the same tactile object classification dataset. Additionally, we propose a way to create more training examples from the recorded data. The results show that the proposed method improves the maximum accuracy from 82.4 \% (BioTac SP fingertips) and 90.7 \% (WTS-FT fingertips) with complete time-series data to about 94 \% for both sensor types.
\end{abstract}

\begin{IEEEkeywords}
3D-CNN, CNN, LSTM, tactile sensing, object classification, BioTac, WTS-FT
\end{IEEEkeywords}

\section{Introduction} \label{ch1:intoduction}
The sense of touch is critical for exploring the properties of objects and their classification. Unlike vision which is subject to many limitations, such as illumination, background noise, occlusion etc., tactile information allows one to directly infer material properties of the object touched. Thus, it is crucial for modern robotics as the field advances towards autonomous grasping, slippage detection, object detection, and dexterous manipulation. Currently employed visual object recognition strategies are challenged, because objects may have similar shape and appearance but different material composition. In contrast, tactile classification strategies incorporating features of the objects shape and material properties have proven to be promising for such settings. \\

\begin{figure}[ht]
        \centering
        \includegraphics[width=0.5\textwidth]{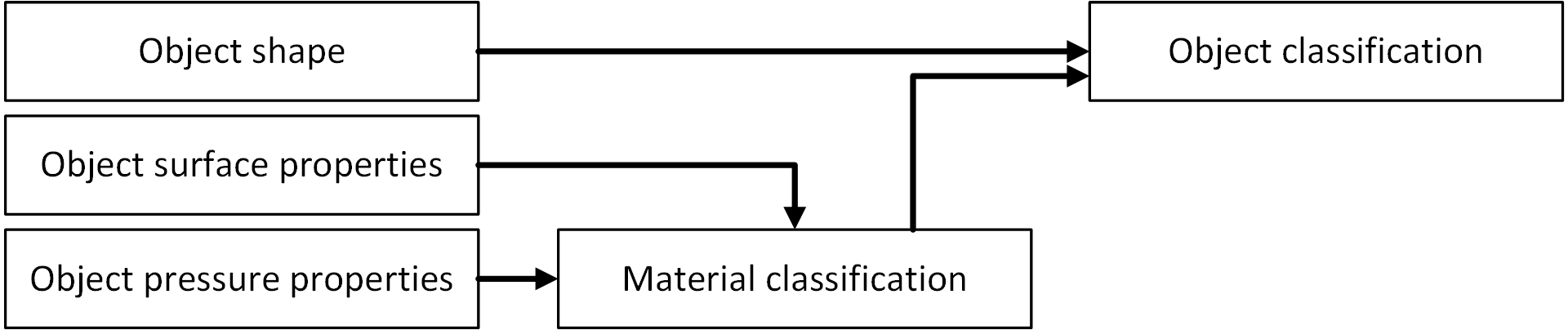}
        \caption{Illustration of haptic object classification steps}
        \label{fig:structure}
    \end{figure}

Haptic object classification systems use the material properties of the objects as well as their shape in the classification step \cite{li_sensing_2013,eguiluz_multi-modal_2016,Kerr2013,Jamali2010,Kerr2014,Kerr2014a,Tanaka2017} as shown in Fig. \ref{fig:structure}. The topic of haptic object classification has seen a wide variety of different data processing and classification approaches. At the same time, it is closely related to other fields in robotics research. Autonomous robotic systems for manipulation are often limited by the use of visual exploration only, thus restricting the identification of the nature of the object and impacting upon the choice of optimal grasps. Haptic exploration at the time of grasp can therefore contribute vital information. Current research features a variety of different approaches for haptic object classification. However, as stated in \cite{baishya_robust_2016}, the variety of test sets and the lack of a standard robot-hand/sensor configuration make it difficult to perform a meaningful comparison of an algorithm's performance.

The trend towards using deep learning in robotics  \cite{schmitz_tactile_2014} has indicated the superiority of these models over traditional machine learning strategies for haptic object classification. This paper contributes a comparison of two different tactile sensors on the same dataset performing the classification of objects through a unimodal haptic approach using a time series of tactile images. It furthermore investigates the performance of 2D-CNN, 3D-CNN and LSTM artificial neural networks used in an end-to-end manner to find an optimal neural network architecture for haptic object recognition. Unlike many publications, the approach utilises spatio-temporal data, facilitating high-resolution tactile sensor data, preserved in its three-dimensional nature. Another novelty to this field is using a three-dimensional CNN for this task and two sensor solutions on the same object set. As opposed to the two-stage approach in \cite{schneider_object_2009,drimus_design_2014}, neural networks feature the ability to learn the relevant features from the data and classify the object in the same training process. Current state-of-the-art papers mostly create hand-crafted features \cite{drimus_design_2014, liu_tactile_2012} or use results of unsupervised models \cite{schneider_object_2009,navarro_haptic_2012} which are then fed to a classification algorithm.
\\
 
The structure of the paper is as follows: a literature review is briefly discussed in section \ref{ch2:lit_review}, the methodology is reported in section \ref{ch3:methodology}, results are presented and analysed in section \ref{ch4:results} and conclusions and future work are discussed in section \ref{ch5:conclusion}.
\section{Background and Related Works} \label{ch2:lit_review}

In this section, two essential aspects of ML algorithms, namely feature extraction and classification, are discussed. Current state-of-the-art one-step approaches employed for the haptic object are reviewed, highlighting the advantages of a \emph{one-step} approach in classification followed by challenges in comparing and reproducing results due to lack of uniformity and consistency in datasets, manipulators, robots and sensors. 

\emph{Two-stage classifiers} includes those object classification solutions which pursue an approach consisting of a first step to extract features from the data on the object in the training period and then uses these features to classify the object in a second step, whilst keeping these two as separate entities. In general, two main approaches to generating features from haptic data are normally used: manual feature design and automatic feature extraction using machine learning methods. Manual feature extraction reduces the volume of the raw data recorded by applying processing rules on the recorded data, which are predefined by the authors \cite{drimus_design_2014,baishya_robust_2016}. Automatic feature extraction uses unsupervised machine learning to pre-process the raw data received from the sensors \cite{navarro_haptic_2012,schneider_object_2009,liu_tactile_2012}.

\emph{Classification}: The data yielded from the feature extraction task constitutes a labelled data-set. Consequently, the remaining classification task is performed by applying a supervised learning algorithm. The most popular classifiers in this field are $k$-nearest-neighbours ($k$-NN) and multi-class support vector machines (SVM). Those have the advantage of being easy to set up and computationally inexpensive when performing classification tasks. Computational complexity substantially depends on the value for $k$ with the k-NN algorithm and on the chosen kernel for the SVM. Moreover, random forest approaches are used as well. Those have the advantage that the most relevant features can be extracted, which might be advantageous for improving the robotic setup. Some papers such as  \cite{schneider_object_2009} use an approach that returns a histogram of different feature comparison algorithms.

\subsection{One-step classification/DeepL classification}

One-step classification in the context of this paper means that the system which performs the classification task is trained and setup as one single entity. Hence feature extraction, representation and subsequent classification are all performed within the same machine learning component. Data pre-processing such as normalisation, Principle Component Analysis (PCA) or similar are not considered a step in this definition and may be performed before training and evaluation. Because many recent publications opted for deep learning methods in their one-step solutions. the term \emph{DeepL classification} will be used synonymously in this paper.\\

Schmitz et al. \cite{schmitz_tactile_2014} claimed to present the first paper to use a deep learning approach in haptic object detection and utilised an architecture of stacked denoising autoencoder (DAE) layers which were pre-trained unsupervised using dropout for further regularisation of their network. They performed extensive hyperparameter optimisation (learning rate, number of hidden layers, dropout rate, activation function, weight penalties) to optimise their solution's performance.
The authors used a TWENDY-ONE robotic platform \footnote{Available online, \protect\url{http://www.twendyone.com}, last accessed 21/4/2021} equipped with a four-fingered hand composed of $241$ haptic pixels distributed over its surface, as well as information on the motor positions to collect data. 

Despite this being a multimodal approach, there is no difficulty in representing the data haptic and other information in the same data vector. The authors created a 20 household objects test set with five similarly shaped bottles, almost certainly making it the most challenging data-set at the time. As a comparison for their proposed classifier, they also tested a shallow artificial neural network with prepossessing through 90\% variance PCA\footnote{Dimensionality only reduced by 8, showing almost no redundancy in the input data} which yielded only $49.3 \%$ accuracy on the test set. The deep DAE network achieved $88.4 \%$ in its best configuration, showing a clear superiority with a DeepL approach.

In contrast, Liarokapis et al. \cite{liarokapis_unplanned_2015} utilised a random forest classifier for haptic object classification, assuming properties such as faster processing speed, inherent feature variable importance calculation and high accuracy of the random forests classifier. They used a two-finger under-actuated gripper with 8 tactile sensors on each finger and recorded 4 motor positions. As data were acquired at two points in time per measurement, a 36-dimensional feature vector was created and used directly for the detection process.

Data collected from 10 household items objects were recorded at different angles and then split into two sets. One contained only object positions called Constrained Orientation and one Free Orientation set, which has no limitations on its angles at which the object was presented to the robot. For the former, a 100 tree random forest yielded an accuracy of $100 \%$, whereas, for the latter, this dropped to $94.3 \%$. Compared with SVMs and shallow NNs, the chosen method is more accurate even for 10 random tree forests. \\

An analysis of the relevant features revealed that 12 of the 16 tactile values could be neglected with only a minor decrease in accuracy ($< 2 \%$). The relevance of shape-related data from the motors appeared to be more important than haptic data in the given scenario. This is in contradiction to \cite{schmitz_tactile_2014}. Considering the much higher commonality of the data in Schmitz et al., this suggests that haptic data needs a certain resolution before its relevance for classification increases.

More recent work by Gao et al. \cite{gao_deep_2016} compared the use of CNNs and LSTMs for assigning property labels to the objects. Furthermore, they also incorporated a visual means of describing the object allowing for a comparison. For the tactile component, a two-finger setup was used. Each finger was equipped with a sensor that delivers 19 readings related to local pressure and low-frequency fluid pressure, high-frequency fluid vibrations, core temperature and core temperature change. 

As the dimensions of the tactile data were reduced to 4 via PCA, the system delivered a 32-signal feature vector per time step. This was either passed to the LSTM as a time-sequence of vectors or as a 2D matrix with features as the first and time as the second dimension. As the task was not to identify an object but rather to assign an unspecified number of 20 possible labels to it, the paper used the \emph{Area under curve} metric to calculate the system's performance. Accuracy can therefore not be compared with the other papers discussed in this section. Gao et al.\cite{gao_deep_2016} achieved an AUC of $72.1 \%$ for the LSTM and $83.2 \%$ for the CNN. The CNN solution was able to outperform every visual descriptor based on the Image-GoogleNet architecture; the results of the LSTM were significantly inferior. 

Table \ref{tab:main_characteristics} summarises a performance comparison of the prominent research works discussed above.

\begin{table}

\caption{Main characteristics of papers employing a single-step classification approach.}
         \resizebox{\columnwidth}{!}{%
        \begin{tabular}{|l|l|l|l|l|l|l}
        	\toprule 
        	Paper & Archi- & Classes & Data & Exploratory & Accuracy \\ 
             & tecture &  & & movement &  \\ 
        	\bottomrule
        	\textbf{Schmitz} & stacked & 20 objects & 241 tactile & simple & $88,3 \%$  \\ 
        	\textbf{et al.}\cite{schmitz_tactile_2014} & DAE & household & 321 feat. & grasp &  \\ 
        	& & & numb. grasps & & \\
        	\hline 
        	\textbf{Liaroka-}& Random & 10 objects & 16 touch & simple & $100 \%$ constr. \\ 
        	\textbf{pis et al.} \cite{liarokapis_unplanned_2015}& forests & household & 2 motor & grasp, & pos. \\ 
        	& & & & 2 time & $94.3 \%$ free pos.\\
        	\hline 
        	\textbf{Gao et al.}& CNN \& & 20 & 32 features & hold, squeeze & CNN out- \\ 
        	\cite{gao_deep_2016} & LSTM & adjectives & 2 sensors, time & slide s/f & performs \\ 
        	& & & & & LSTM \\
        	\hline 
        	\textbf{Baishya \&} & CNN & 6 tubes & 4x4 touch & sliding & $92.3 \%$ 1 slide \\ 
        	\textbf{Baeuml} \cite{baishya_robust_2016} &  & diff. material & 1500 time &  & $97.5 \%$ 3 slides \\ 
        	\bottomrule
        \end{tabular} 
  }     
        \label{tab:main_characteristics}
    \end{table}

\subsection{Comparability of data processing solutions}
The field of haptic object classification has seen various proposed systems that rely on very different setups of sensors (high/low spatial resolution touch sensors, one/multiple sensors) and test objects (geometrical objects, household objects). In general, it is difficult to compare and reproduce the results, as the approaches use different and very often unique hardware. They lack generality or have not been tested across several platforms. Some methods consider using only one sensor type, which means either tactile or kinesthetic modality \cite{navarro_haptic_2012}.
    
Therefore, accurate comparison of algorithms can only be performed using the same environment and hardware. For example, \cite{baishya_robust_2016} provide a comprehensive review of classical approaches using high and low dimensional features as well as 2D-CNN neuronal networks. However, their data set is basically the same shape although multi-material, which means it cannot be compared with, for example, \cite{gao_deep_2016} who also uses LSTMs as their ML approach but exploit household objects, which possess a different range of material and shape properties. This paper focuses on the same ML approaches using a single dataset and robot hand but with two different sensors, allowing for convenient and robust comparison.

\section{Methodology} \label{ch3:methodology}
    \subsection{Setup}
        The mechanical setup for the haptic object classification comprises a human-inspired robotic hand and high-resolution tactile sensors. To gain insight into the influence of the sensor type on our recognition rate, two distinct types of sensors were used: \emph{(1.)} The first is the curved BioTac SP sensor which features $24$ taxels (tactile pixels), not aligned in a grid but in a bionic inspired layout. Furthermore, this sensor provides temperature and vibration data which have been facilitated as means of material classification in \cite{eguiluz_multi-modal_2016}. \emph{(2.)} The second device is the WTS-FT sensor by Weiss robotics, which exploits a flat surface with a slight bend towards the tip and a resolution of $4 \times 8$ taxels. These sensors were mounted to the thumb, the index and the ring finger of an AR-10 humanoid robotic hand which provided the mechanical actuation.
        \begin{figure}
            \center
            \includegraphics[width=0.3\textwidth]{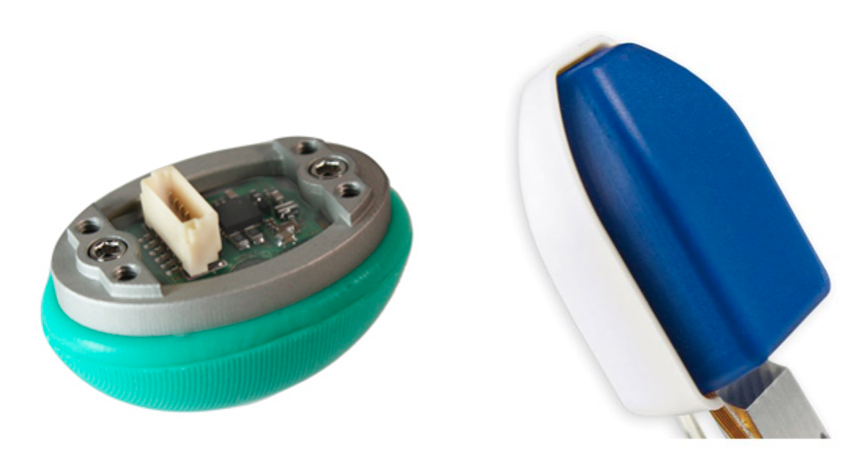}
            \caption{\emph{Left:} The BioTac-SP bionic inspired tactile sensor; \emph{Right:} The WTS-FT tactile sensor}
            \label{fig:sensor_img}
        \end{figure}
    
    \subsection{Objects}
        A haptic object classification system is expected to extract as much information from a single grasp as possible. Therefore, a set of objects was composed to evaluate the method’s capabilities to infer object shapes and detect basic material properties of the grasped objects. Practically, this meant including a variety of elementary shapes and items with different stiffness and elasticity but similar silhouettes. The system must differentiate between three cylindrical objects comparable in terms of size but distinct in their softness. Additional to the fundamental shapes of a cylinder, a sphere and a box, two special items were designed to pose a greater challenge to the classification algorithms: A triangular prism and an Icosahedron which both possess the property to present a varying combination of surfaces, edges and angles to the tactile sensors depending on the item's alignment in space. Classifying these objects requires the model to extract the dynamic behaviour of the object during the grasp.
        \begin{figure}
            \center
            \includegraphics[width=0.5\textwidth]{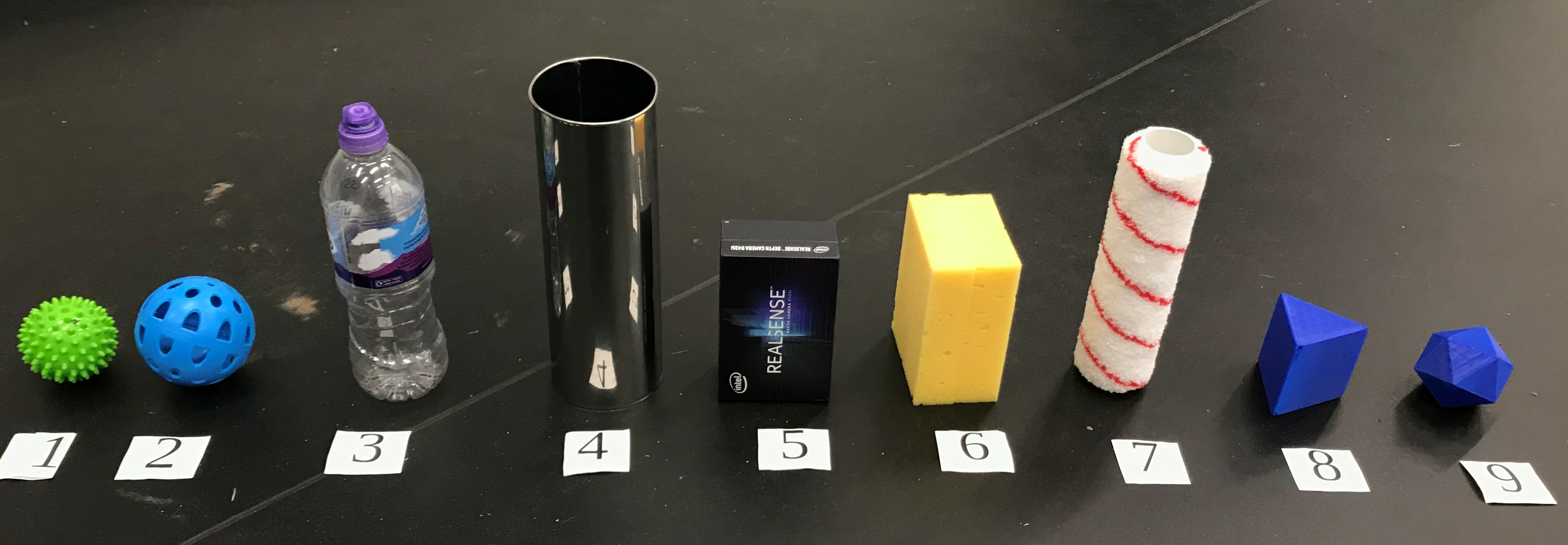}
            \caption{The set of objects which were to be classified by a single grasp: (1) Spiky rubber ball; (2) Solid light plastic ball; (3) Empty water bottle (sealed); (4) Thin metal pipe; (5) Solid cardboard box; (6) Box-shaped sponge; (7) Soft cylindrical painter's roller; (8) 3D-printed triangular prism; (9) 3D-printed icosahedron}
            \label{fig:grasp_objects}
        \end{figure}
    
    \subsection{Data recording}
        To keep the data acquisition as close to a real use case as possible, while maintaining appropriate recording intervals, the AR-10 hand was mounted on a desk while manually feeding the object to the hand. On presentation of the object, the hand closed its fingers/thumb. Data recording commenced as soon as the following condition to detect a grasp was met: the haptic sensor of the thumb and at least one of the two other fingers recognises pressure on its taxels. If this criterion was not constantly met within the first second of the recording, the data was ignored, and the acquisition started over again. Approximately $\Delta t = \SI{6}{s}$ long periods of recording were performed while the grasping programme executed finger movement to secure its grasp. For the BioTac this resulted in 63 samples per recording and about 200 for the WTS-FT. However, for better comparability, the WTS-FT data were downsampled to 63 samples per time series. \\

        Each item in the object set was grasped 40 times with varying angles of attack for the fingers. Passing the object by hand further contributes to introducing data variance as it is practically impossible to recreate precisely the same handover condition. This mimics the case of human-robot handover in an industrial setting.
    
    \subsection{Data pre-processing}
        \paragraph{Normalisation \& Filtering}
        The recorded time series undergo a data prepossessing routine before passing on to the machine learning algorithm. Firstly, the taxel values from both the BioTac SP and WTS-FT sensors are normalised by either of the two following methods: Map the interval from zero to the maximal value to 0…1 (from now on referred to as \emph{scaled}) or map the values in a way that they are arranged like a standard Gaussian distribution ($\mu = 0$, $\sigma = 1$) (from now on referred to as \emph{normalised}). The normalisation technique used is highlighted in the results tables \ref{tab:wts_table} \& \ref{tab:biotac_table} in the "norm" column. Subsequently, the data is reshaped in a suitable way for the ML method employed. Hence, all the values from the three sensors are concatenated to one vector per time-step for LSTM training. For the CNN models, the 2D nature of the sensor output is preserved, and the three tactile images are joined together in the order: thumb, index, ring finger.\\

        As the taxels of the BioTac sensors are not aligned with a grid, these measurements need to undergo an additional pre-processing step when used for a CNN model. The tactile dots are mapped onto a matrix in a way that preserves their actual distance relationships, similar to the method in Zapata-Impata et al. \cite{zapata-impata_non-matrix_2018}. As proposed in \cite{zapata-impata_non-matrix_2018} there is also the option to filter over the yielded matrix to blur the position of the taxels and create the impression of a tactile image instead of a sparse matrix with some non-zero elements which was not adopted in our implementation.

\begin{figure}
            \center
            \includegraphics[width=0.25\textwidth]{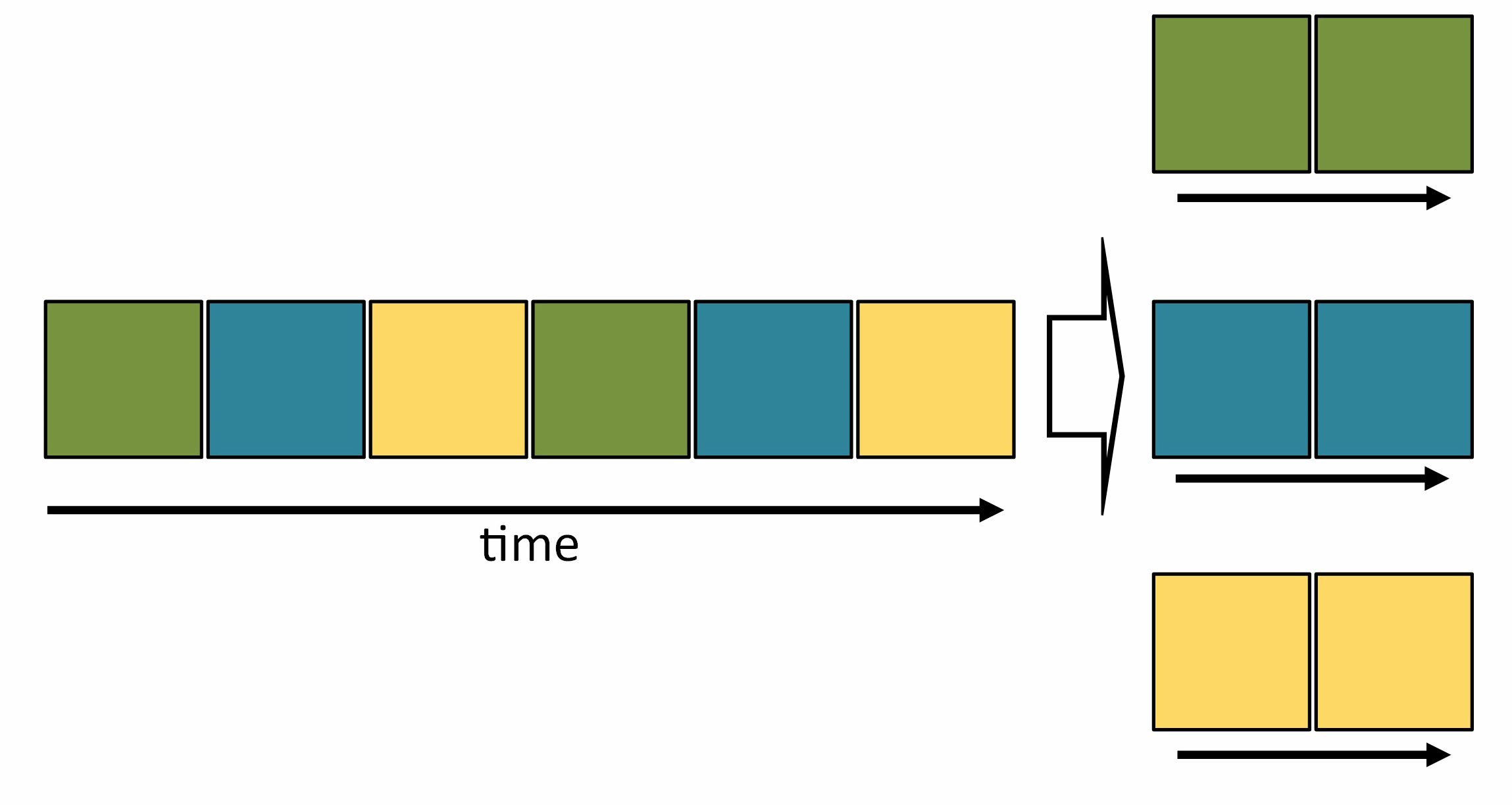}
            \caption{The way time-series are split to extend the available data for training}
            \label{fig:data_augm}
        \end{figure}

        \paragraph{Data splitting}
        Recording data for a haptic dataset is cumbersome and time-consuming. There are a limited number of ways to speed up the data capturing process as it always involves a physical grasp.  Because robotic grasping is a relatively slow process captured multiple times per second, it is assumed that a much lower sampling rate might actually be sufficient to build a working classifier. Therefore, the data is down-sampled but without omitting any data in this process. As depicted in fig. \ref{fig:data_augm}, the data points which would be neglected by the downsampling are reassigned to newly created time-series.\cite{depp_haptic} Consequently, the redundancy in the data is being used to boost the dataset size to provide more training samples. However, these examples are likely to be very much akin to each other. Hence, the split between training, testing and validation dataset needs to occur before this method is applied. Otherwise, overfitting inevitably occurs due to high similarities between both datasets, which leads to unrealistically high validation accuracy.
    \subsection{Convolutional network classifier}
        Convolutional Neural Networks (CNN) are rapidly becoming the state-of-the-art method for evaluating images and comparable data. Therefore, they are an obvious choice for the tactile pictures that were recorded. Despite that, the existence of a time dimension distinguishes it from a conventional image classification. In \cite{drimus_classification_2011}, the authors suggest that transient processes during the grasp may account for better classification potential, and accordingly, we considered this aspect in the design of our network. One possibility is to include the extra time-steps as additional channels for the CNN’s input. In this case, the network features up to 63 individual input channels with their own set of convolutional kernels. Another way to account for time-dependent processes is to employ the 3D version of a CNN, which features three-dimensional tensors instead of matrices as kernels. This design choice also creates the opportunity to investigate whether kernels with significant extensions in the space or the time-dimension lead to better performing models. Assuming that bigger kernels can capture more complex patterns, this might indicate whether the time or the spatial resolution of tactile sensing data contribute more substantially to robust object classification.
        
\begin{figure}[htb!]
			\centering
			\subfloat[]{{\includegraphics[width=0.15\textwidth]{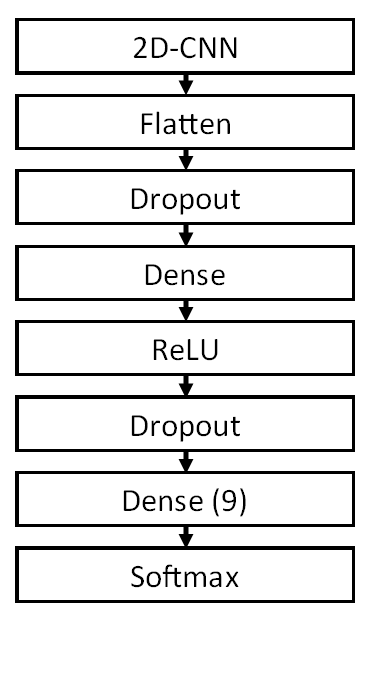} }}
			\qquad
			\subfloat[]{{\includegraphics[width=0.15\textwidth]{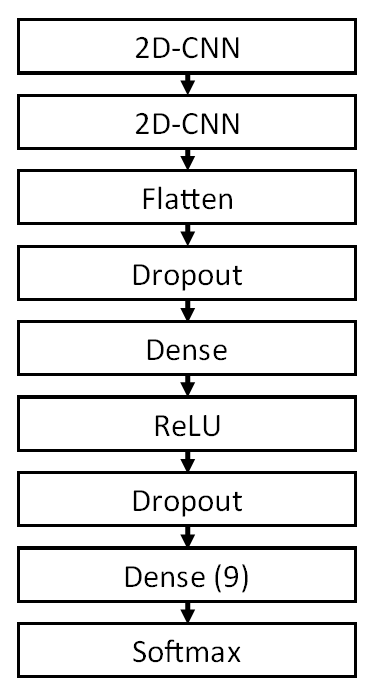} }}
			\qquad
			\subfloat[]{{\includegraphics[width=0.15\textwidth]{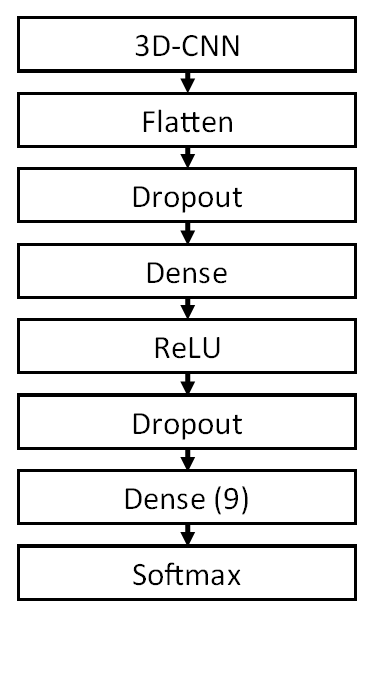} }}
			\qquad
			\subfloat[]{{\includegraphics[width=0.15\textwidth]{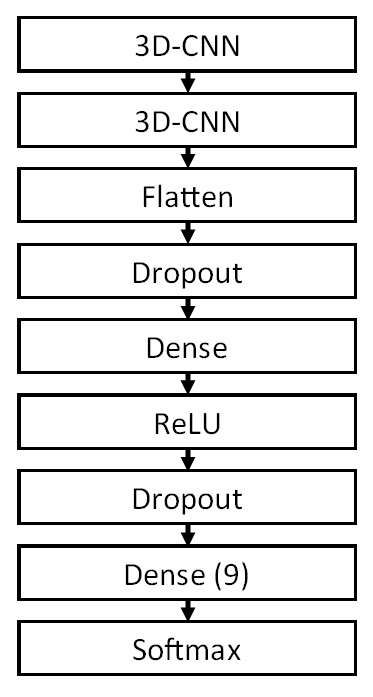} }}
			\caption{CNN architectures used: (a) 2D-CNN-1L (b) 2D-CNN-2L (c) 3D-CNN-1L (d) 3D-CNN-2L}
			\label{fig:cnn_arch}
		\end{figure}

\begin{figure}[t]
            \centering
            \includegraphics[width=0.15\textwidth]{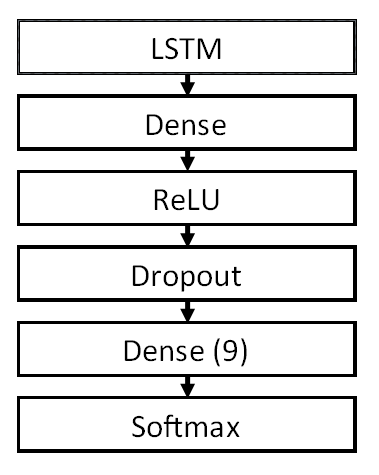}
            \caption{LSTM architecture used}
            \label{fig:lstm_arch}
        \end{figure}

        A number of ML approaches were employed to analyse the tactile data, namely four variants of CNN networks and one LSTM network - as summarised in figures \ref{fig:cnn_arch} and \ref{fig:lstm_arch}.
        Both the 2D and 3D CNN variants were used in a single and two layer configuration to assess an appropriate network depth for the classification task as shown in fig. \ref{fig:cnn_arch}. Due to overfitting being considered a potential concern with the limited data available, two dropout layers were integrated to facilitate better model generalisation in the training process. Also, a horizontal ($\pm 1$ taxels) and vertical shift (up to $\pm 2$ taxels)  were used as a data augmentation method during model training. Recognising the small resolution of the tactile image, more elaborate methods of image augmentation as for example, in Shoten and Khoshgoftaar \cite{shorten_survey_2019} were not employed.
    
    \subsection{Recurrent network classifier}
        LSTMs have proven to be a powerful tool for machine learning on time-series data. However, their structure does not incorporate multi-dimensional input vectors. 

        Therefore, the tactile data captured at each time step was concatenated to one vector comprising a series of 1D arrays. The LSTM layer was set up to only pass output to the subsequent dense layer after processing the whole time series. Dropout was activated for the LSTM nodes and recurrent dropout to the recurrent inputs of the layer to mitigate overfitting. The dense layer was also combined with a dropout layer. These approaches against overfitting are more crucial for training the LSTM than for the CNN architectures. The methods for data augmentation used on the 2D data cannot be facilitated with the 1D data representation. Furthermore, it was decided to evaluate only a single LSTM layer architecture (fig. 6) because a two-layer solution with the same range of nodes as for the CNN would yield a much bigger model for an already overfitting-prone dataset.
    \subsection{Network training \& Optimisation}
        All networks were trained using the \emph{RMSprop} optimiser and the ReLU activation function. The parameters of the optimiser and the number of nodes for the CNN/LSTM layer(s) and the dense layer were assigned through hyperparameter search using the Talos library \cite{talos_sw}. A similar approach was used for the dropout ratios. This strategy ensured that all network configurations were judged by their actual model performance and not by the quality of manually selected hyperparameters.

\section{Results} \label{ch4:results}
    \subsection{Accuracy}
\begin{figure}[t]
            \centering
            \includegraphics[width=0.45\textwidth]{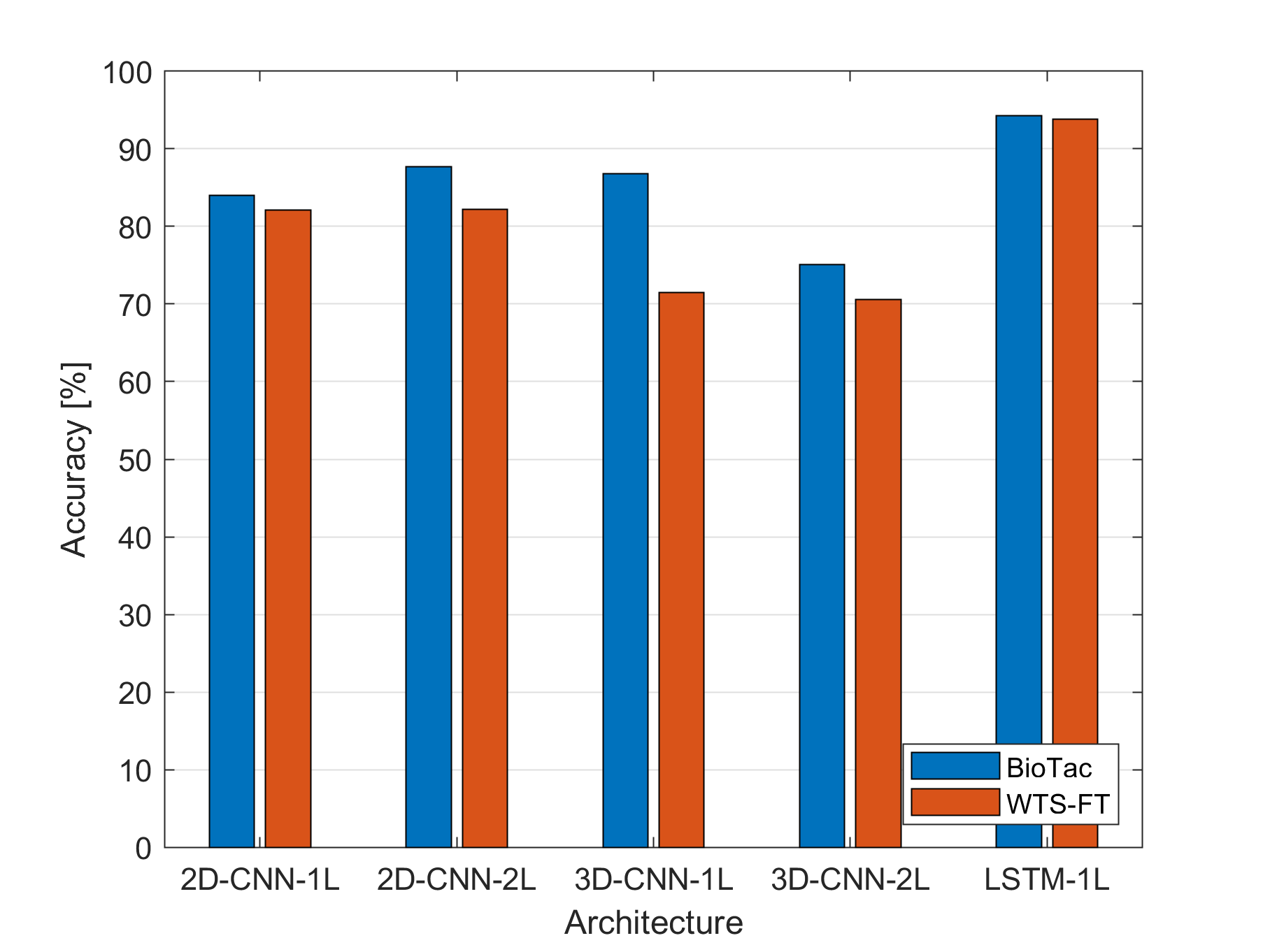}
            \caption{Maximum accuracy per architecture and sensor}
            \label{fig:bar_acc}
        \end{figure}

        The main objective of this paper was to evaluate the performance of different neural network-based models on an object recognition task, comparing results from data acquired using two robotic tactile sensors. From fig. \ref{fig:bar_acc}, it may be observed that for the recorded dataset on the selected group of objects, the LSTM based network performs best for both sensor types with a best accuracy of 94.3 \% of the objects being classified correctly. (BioTac,Table~\ref{tab:biotac_table}) and 93.8 \% (WTS, Table~\ref{tab:wts_table}) respectively. For the CNN architectures, it may be seen the performance of the BioTac sensors proved to be slightly better than that of the WTS-FT sensors. 
        
        In terms of CNN based classifiers, the 2D-CNN models were generally slightly more accurate than the 3D-CNN versions. Nevertheless, the performance results were close between the best 3D-CNN with one CNN layer and the two 2D-CNN model (comp. fig. \ref{fig:bar_acc}). However, for robotic systems, the computational effort needs to be taken into account. Thus, 3D-CNN networks may still be considered, as the 2D-CNN models contain significantly more parameters causing higher computational cost. That is due to having a set of filters for each channel instead of the single set of filters for the 3D-CNN. 

        The confusion matrices depicted in fig. \ref{fig:confusion_matr} provide information on the misclassifications that occurred with the best CNN and LSTM model on the two sensors. The CNN architecture using the BioTac sensors misclassifies the two balls in the dataset as an icosahedron akin to a spherical object. However, the CNN-based classifier's major flaw with WTS-FT sensors appears to be the distinction between the plastic bottle and the metal cylinder – a mistake seen for the LSTM models with the same sensor type as well. Therefore it can be inferred that this was not caused by inherent insufficiency in the model but rather by the sensor design itself. Finally, the LSTM model with BioTac data exhibits a different kind of model error by wrongly mapping the spiky ball's data to the classes of the prism or icosahedron.
        \begin{table}[h]
            \centering
            \caption{BioTac validation accuracy}
            \begin{tabular}{|p{1.0cm}|p{1.0cm}|p{0.8cm}|p{1.0cm}|p{1.0cm}|}
                \toprule
            	\multicolumn{5}{c}{\textbf{2D-CNN}}\\
            	\bottomrule
            	Number of layers  & split ratio & time-series length & norma-lisation & validation acc.  \\
            	\toprule
            	1 & 1 & 63 & scaled, filtered & 62.0 \% \\ 
            	1 & 3 & 21 & scaled, fit. & 70.1 \% \\ 
            	1 & 5 & 13 & scaled, fit. & 76.6 \% \\ 
            	1 & 7 & 9 & scaled, fit. & 70.9 \% \\ 
            	1 & 1 & 63 & scaled & 72.2 \% \\ 
            	1 & 3 & 21 & scaled & 72.5 \% \\ 
            	1 & 5 & 13 & scaled & 82.2 \% \\ 
            	1 & 7 & 9 & scaled & 84.0 \% \\ 
            	\midrule
            	2 & 1 & 63 & scaled & 67.6 \% \\ 
            	2 & 3 & 21 & scaled &  80.2 \% \\ 
            	2 & 5 & 13 & scaled & 82.0 \% \\ 
            	2 & 7 & 9 & scaled & 87.7 \% \\ 
            	\toprule
            	\multicolumn{5}{c}{\textbf{3D-CNN}}\\
            	\bottomrule
            	Number of layers  & split ratio & time-series length & norma-lisation & validation acc.  \\
            	\toprule
            	1 & 1 & 63 & scaled & 68.5 \% \\ 
            	1 & 3 & 21 & scaled & 71.6 \% \\ 
            	1 & 5 & 13 & scaled & 78.9 \% \\ 
            	1 & 7 & 9 & scaled & 86.8 \% \\ 
            	\midrule
            	2 & 1 & 63 & scaled & 54.6 \% \\ 
            	2 & 3 & 21 & scaled & 70.0 \% \\ 
            	2 & 5 & 13 & scaled & 69.7 \% \\ 
            	2 & 7 & 9 & scaled & 75.1 \% \\ 
            	\toprule
            	\multicolumn{5}{c}{\textbf{LSTM}}\\
            	\bottomrule
            	Number of layers  & split ratio & time-series length & norma-lisation & validation acc.  \\
            	\toprule
            	1 & 1 & 63 & scaled & 82.4 \% \\ 
            	1 & 3 & 21 & scaled & 90.7 \% \\ 
            	1 & 5 & 13 & scaled & 94.3 \% \\ 
            	1 & 7 & 9 & scaled & 90.0 \% \\ 
            	\bottomrule
            \end{tabular} 
        
            \label{tab:biotac_table}
        \end{table}

        \begin{table}[h]
            \centering
            \caption{WTS-FT validation accuracy}
            \begin{tabular}{|p{1.0cm}|p{1.0cm}|p{1.0cm}|p{1.0cm}|p{1.0cm}|}
            	\toprule
            	\multicolumn{5}{c}{\textbf{2D-CNN}}\\
            	\bottomrule
            	Number of layers  & split ratio & time-series length & norma-lisation & validation acc.  \\
            	\toprule
            	1 & 1 & 63 & norm. & 68.5 \% \\ 
            	1 & 3 & 21 & norm. & 69.8 \% \\ 
            	1 & 5 & 15 & norm. & 67.8 \% \\ 
            	1 & 7 & 9 & norm. & 70.9 \% \\ 
            	1 & 1 & 63 & scaled & 82.1 \% \\ 
            	1 & 3 & 21 & scaled & 80.9 \% \\ 
            	1 & 5 & 15 & scaled & 81.5 \% \\ 
            	1 & 7 & 9 & scaled & 77.9 \% \\ 
            	\midrule
            	2 & 1 & 63 & scaled & 78.7 \% \\ 
            	2 & 3 & 21 & scaled &  77.8 \% \\ 
            	2 & 5 & 15 & scaled & 82.2 \% \\ 
            	2 & 7 & 9 & scaled & 79.2 \% \\ 
            	\toprule
            	\multicolumn{5}{c}{\textbf{3D-CNN}}\\
            	\bottomrule
            	Number of layers  & split ratio & time-series length & norma-lisation & validation acc.  \\
            	\toprule
            	1 & 1 & 63 & scaled & 71.3 \% \\ 
            	1 & 3 & 21 & scaled & 65.4 \% \\ 
            	1 & 5 & 15 & scaled & 71.5 \% \\ 
            	1 & 7 & 9 & scaled & 70.4 \% \\ 
            	\midrule
            	2 & 1 & 63 & scaled & 64.8 \% \\ 
            	2 & 3 & 21 & scaled & 66.7 \% \\ 
            	2 & 5 & 15 & scaled & 70.2 \% \\ 
            	2 & 7 & 9 & scaled & 70.6 \% \\ 
            	\toprule
            	\multicolumn{5}{c}{\textbf{LSTM}}\\
            	\bottomrule
            	Number of layers  & split ratio & time-series length & norma-lisation & validation acc.  \\
            	\toprule
            	1 & 1 & 63 & scaled & 90.7 \% \\ 
            	1 & 3 & 21 & scaled & 93.8 \% \\ 
            	1 & 5 & 15 & scaled & 88.7 \% \\ 
            	1 & 7 & 9 & scaled & 89.9 \% \\ 
            	\bottomrule 
            \end{tabular} 
        
            \label{tab:wts_table}
        \end{table}
        
    \subsection{Performance dependence on time-series}
        In section \ref{ch3:methodology} we suggested a method for splitting data to reduce the effective sampling frequency for the benefit of the dataset size to improve the number of training samples available. 

\begin{figure}[hbt!]
            \centering
            \includegraphics[width=0.45\textwidth]{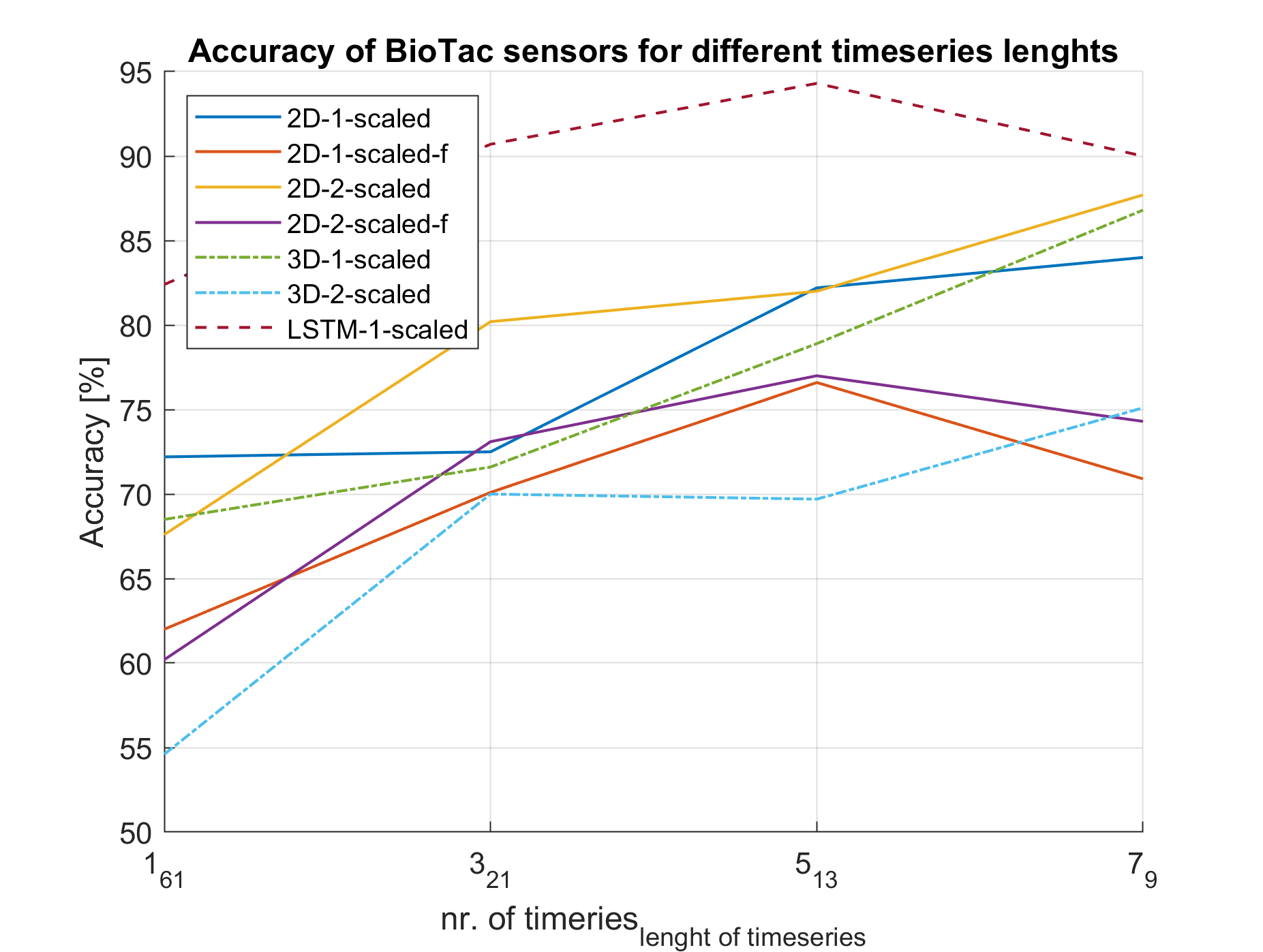}
            \caption{Accuracy on the BioTac-SP sensor over time-series length}
            \label{fig:biotac_acc}
        \end{figure}

\begin{figure}[hbt!]
            \centering
            \includegraphics[width=0.45\textwidth]{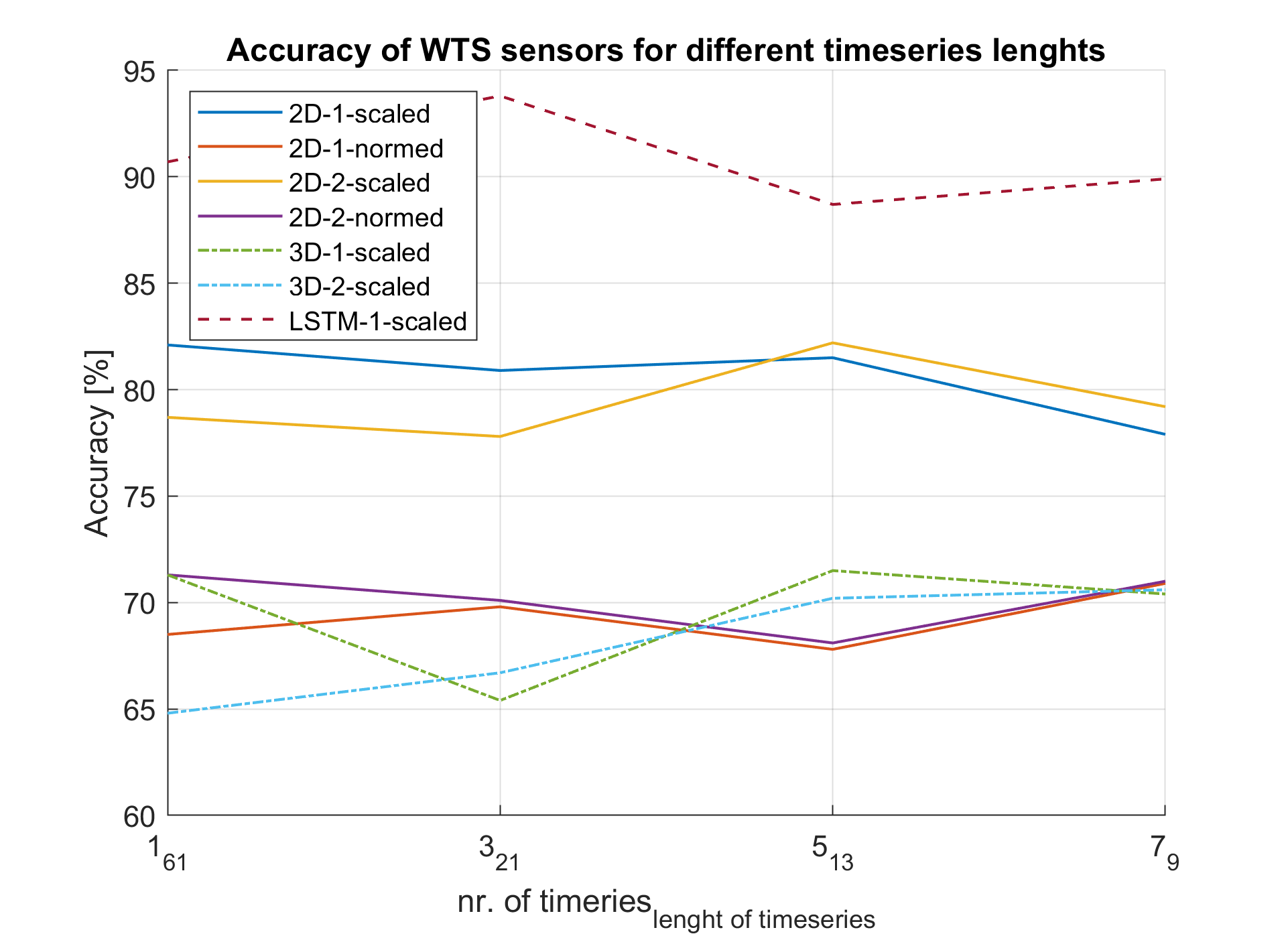}
            \caption{Accuracy on the WTS-FT sensors over time-series length.}
            \label{fig:wts_acc}
        \end{figure}

\begin{figure}[t]
			\centering
			\subfloat[]{{\includegraphics[width=0.25\textwidth]{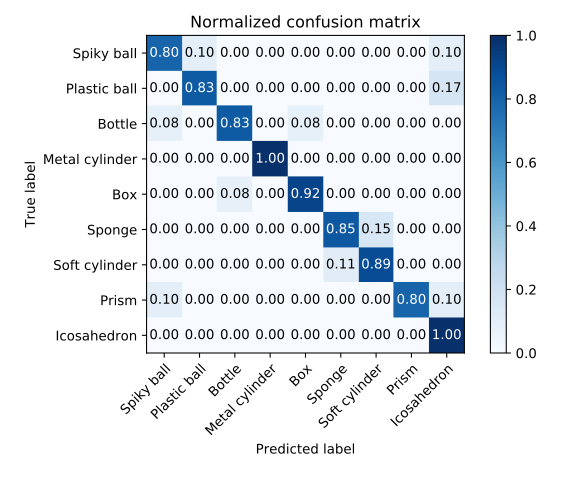} }}
			\qquad
			\subfloat[]{{\includegraphics[width=0.25\textwidth]{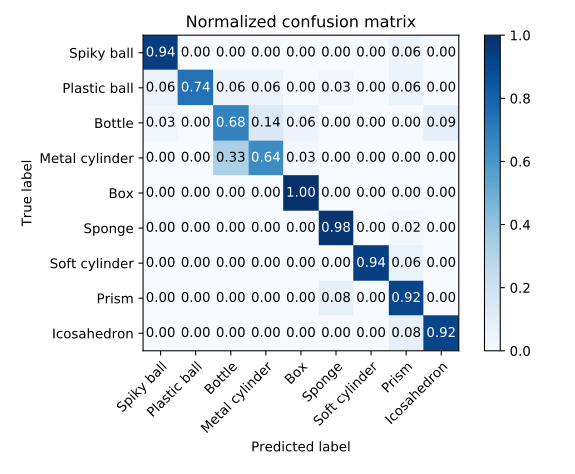} }}
			\qquad
			\subfloat[]{{\includegraphics[width=0.25\textwidth]{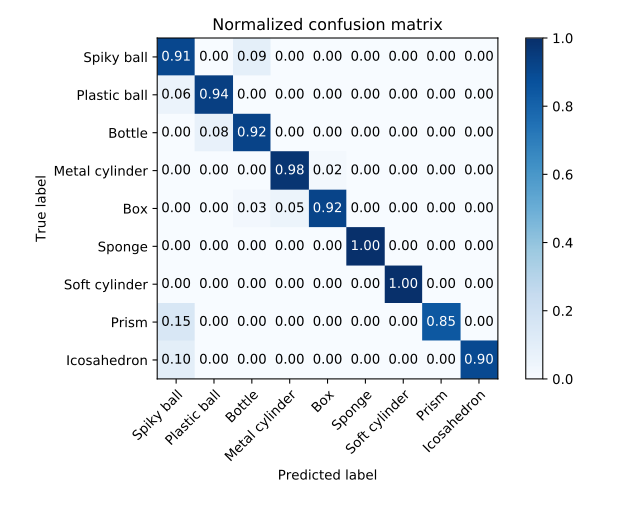} }}
			\qquad
			\subfloat[]{{\includegraphics[width=0.25\textwidth]{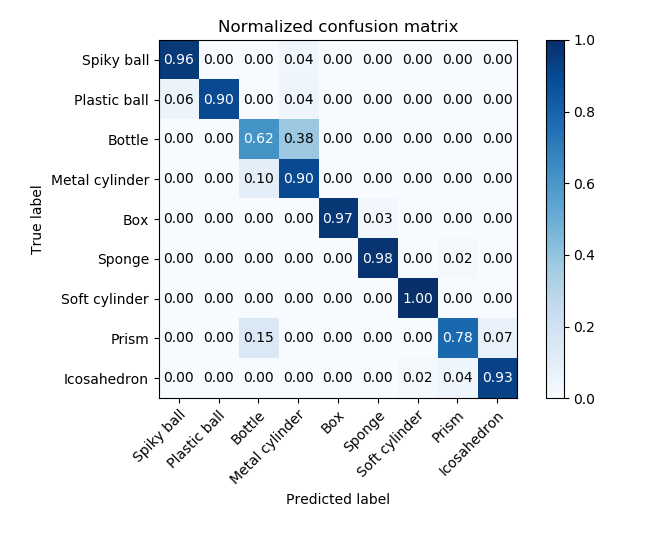} }}
			\caption{Confusion matrices for predictions on the validation dataset: \emph{(a)} CNN - BioTac; \emph{(b)} CNN - WTS-FT; \emph{(c)} LSTM - BioTac; \emph{(d)} LSTM - WTS-FT}
			\label{fig:confusion_matr}
\end{figure}

        As depicted in figures \ref{fig:wts_acc} and \ref{fig:biotac_acc}, the splitting ratio and the normalisation method influenced the maximum achievable accuracy. However, the LSTM architectures benefit from more data and requiring a remaining length of 13 (LSTM for BioTac) or 21 (LSTM for WTS-FT) to work best. The split ratio of 7, which reduces the number of samples per series to 9, appears to be too short to extract the necessary features to tell the objects apart. The data is less conclusive with the CNN architectures as scaled data for the 2D-CNN on WTS-FT sensors appears to be optimally split into 5 subsets. On the contrary, the best CNN models for the BioTac sensor achieve peak performance for the shortest time series length tested. \\
        
        Hyperparameter optimisation for the 3D-CNN included separate parameters for the time and spatial dimension of the convolutional kernels to determine the optimal ratio between these dimensions. However, the results of this procedure have remained inconclusive with regards to this property because the best ten models were found to have quite different combinations of these two parameters.
        
\section{Conclusion and Future Work} \label{ch5:conclusion}
A comprehensive comparison of two tactile sensor technologies using the same robotic hand on a common 9-class dataset was presented in this work. The evaluation was performed using different neural network architectures to exploit the captured information. In that regard, we found LSTM based architectures outperformed CNN models, by a small margin. Furthermore, a method to extend the amount of usable data from a limited number of grasps was also proposed. The proposed method improves the maximum accuracy from 82.4 \% (BioTac) and 90.7 \% (WTS-FT) with full-time series data to about 94 \% for both sensor types.

Additionally, a decline in accuracy for the LSTM models with shorter time series raises the question of obtaining the optimal sampling frequency and grasping recording length. The relevance of the different phases of a grasp for classification performance will be considered in future work. As the confusion matrices for the BioTac have shown that different models exhibit weaknesses on different object classes, ensemble techniques may be considered for a further improvement of accuracy. For deployment purposes, techniques for NN pruning might be helpful to reduce the computational complexity of the proposed classifier.
\flushend
\bibliographystyle{IEEEtran}
\bibliography{IEEEabrv,ref.bib}
\end{document}